\documentclass{article}
\usepackage{graphicx}
\usepackage{caption}
\usepackage[preprint]{corl_2021} 
\usepackage{amsmath}
\usepackage{amsfonts}
\usepackage{booktabs}
\usepackage{siunitx}
\usepackage{times}
\usepackage{textcomp}

\usepackage{amssymb}
\usepackage[normalem]{ulem}
\useunder{\uline}{\ul}{}

\title{Learning by Cheating : An End-to-End Zero Shot Framework for Autonomous Drone Navigation}

\author{
    Praveen Venkatesh, Viraj Shah*, Vrutik Shah*, Yash Kamble*, Joycee Mekie \\
    Indian Institute of Technology, Gandhinagar, India
}

\newcommand\blfootnote[1]{%
  \begingroup
  \renewcommand\thefootnote{}\footnote{#1}%
  \addtocounter{footnote}{-1}%
  \endgroup
}

\begin{document}
\maketitle
\blfootnote{*Equal contribution}

\begin{abstract}
This paper proposes a novel framework for autonomous drone navigation through a cluttered environment. Control policies are learnt in a low-level environment during training and are applied to a complex environment during inference. The controller learnt in the training environment is tricked into believing that the robot is still in the training environment when it is actually navigating in a more complex environment. The framework presented in this paper can be adapted to reuse simple policies in more complex tasks. We also show that the framework can be used as an interpretation tool for reinforcement learning algorithms.

\end{abstract}

\section{Introduction}
Piloting a drone through a crowded environment is a very complex task. Expert human pilots use first-person-view images and are able to navigate through a given environment successfully, even at high speeds. The inherent instability of the quadrotor platform, combined with moving through a cluttered environment, makes navigation safety of utmost importance. Even slight collisions with the environment can cause irrecoverable failures, making the task of navigating even harder. The need for such systems is emphasised in situations such as disasters, construction, surveying, among others. Expert pilots often train for several years before gaining mastery over their technique, leading to high training costs. In this work, we attempt to explore the possibility of autonomously navigating through a cluttered environment safely.

Traditionally, the problem of drone navigation is tackled through separate state estimation and control. Conventional perception methods such as SLAM \cite{SLAM} for state estimation are often unreliable in dynamically changing environments. Deep learning-based perception methods have shown significantly improved robustness compared to traditional methods leading to their widespread adoption \cite{DLPerception}. However, the dual-stage pipeline leads to a decoupling between perception and control. This decoupling may lead to unrecoverable failures in the event of errors in either of the two stages. Hence, there is a need for an end-to-end framework to improve safety and improve system performance.

Our framework takes inspiration from human flying. Pilots maintain a rough estimate of the drone within the world, paying attention to essential aspects of the environment to maintain a safe and smooth flight. These estimates are not always entirely accurate but are merely imprecise estimates of potentially safe vs unsafe areas. We propose a novel end-to-end framework based on this idea to learn control policies based on representations in one environment that are later transferred to different environments. This is where the learnt policy network is \textit{cheated} into believing that it still exists in the environment where it was trained, when it is operating on a \textit{real} environment.

\section{Our Approach}


\begin{figure}[!htb]
  \begin{minipage}{0.49\textwidth}
     \centering
     \includegraphics[width=\linewidth, height=3.25cm]{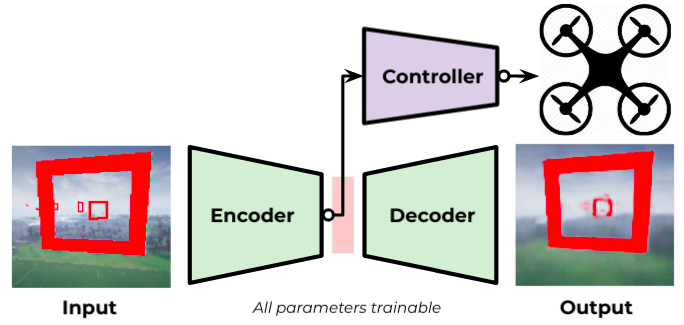}
     \caption{Perception - Variational Autoencoder compresses observed inputs into a low dimensional space on which a control policy is learnt}\label{fig:perception}
  \end{minipage}\hfill
  \begin{minipage}{0.49\textwidth}
     \centering
     \includegraphics[width=\linewidth, height=3.25cm]{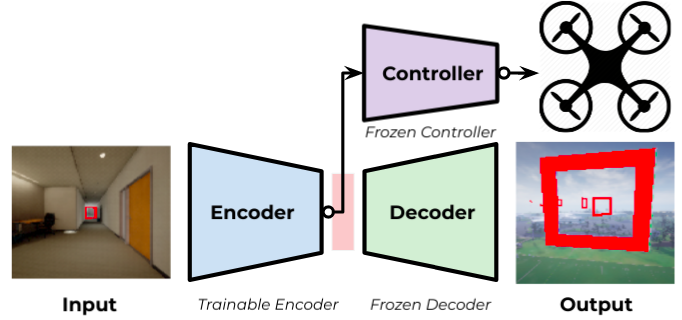}
     \caption{Cheating the network - We replace the encoding stage with a new trainable encoder that learns to map inputs from observed space to the fake environment.}\label{fig:cheating}
  \end{minipage}
\end{figure}


\begin{figure}
    \centering
    \includegraphics[width = \linewidth]{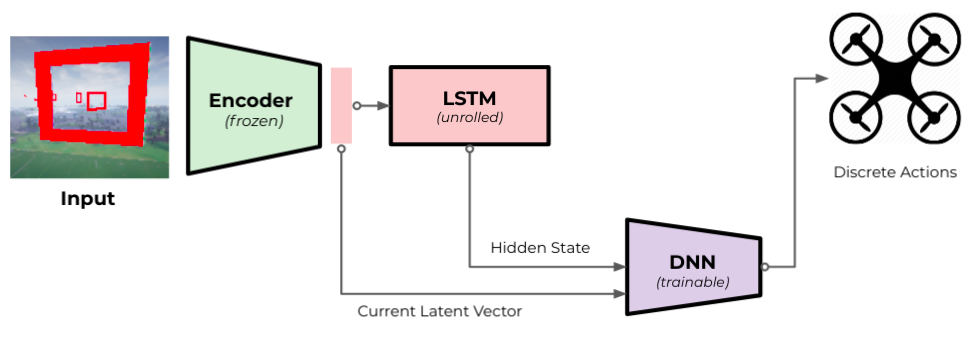}
    \caption{Controller block - A 3 layer MLP is provided with concatenated inputs of 1) Sample from the latent space of the encoder 2) Hidden state of an unrolled LSTM}
    \label{fig:controller}
\end{figure}

In this work, we present a novel zero shot, end-to-end framework for drone navigation. We formulate the problem as a mapping from $\mathrm{R} ^{H\times W\times 3} \rightarrow \mathrm{R} ^ 4$, where the inputs are a three-channel front-facing RGB image, and the outputs are three-dimensional velocity commands and the desired yaw rate.

We divide our approach into three steps: 1) Learning a compressed representation of a simulated environment (referred to as \textit{fake} from here on) 2) Learning a control policy based on compressed representations \& 3) Cheating the learnt controller by learning to predict the compressed features on a different \textit{real} environment. By cheating the learnt controller, we essentially force the network to \textit{believe} that it is still navigating through the \textit{fake} environment, whereas in reality, it is navigating through a \textit{real} environment.
Each of the steps is trained independently, freezing appropriate weights as explained in the following subsections.

\subsection{Learning Compressed Representations}

Representation learning for robotic applications has gained increasing interest in recent years to reduce the dimensionality of a given problem robustly. In this work, we utilize AirSim \cite{airsim}, a high fidelity drone-based environment, to train our representations. We spawn rectangular gates procedurally (as seen in fig \ref{fig:perception}) and use segmented images of these gates to learn a low dimensional representation from RGB images. This low dimensional representation is learnt as a smooth $k$-dimensional space using a convolutional variational auto-encoder(VAE).
i.e., We learn a $\mathrm{R} ^{H\times W\times 3} \rightarrow \mathrm{R} ^ k$ mapping from the input RGB images. This forms our compressed representation space on which a control policy is later learnt. A block diagram of the network can be seen in fig \ref{fig:perception}. The VAE is trained separately on the \textit{fake} environment, and all parameters are trainable. 

\begin{figure}[h!]
    \centering
    \includegraphics[width = \linewidth]{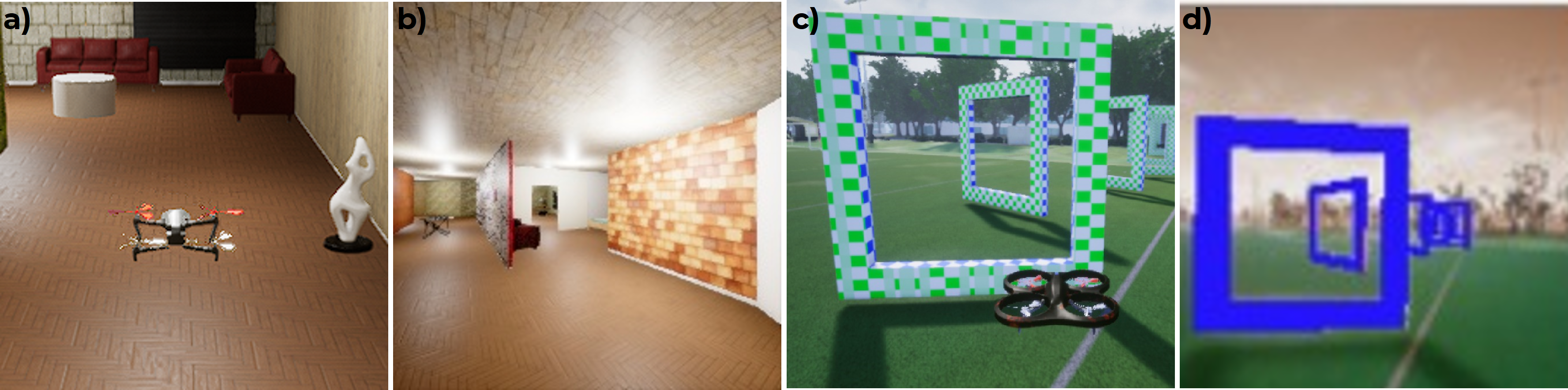}
    \caption{a) \textit{Real} environment to which the policy is transferred. b) FPV view of the drone in the \textit{real} environment. c) \textit{Fake} environment on which the policy is learnt. d) FPV view of the drone in the \textit{fake} environment}
    \label{fig:cheatimg}
\end{figure}

\subsection{Learning a Control Policy}

A diagram of the control network can be seen in fig \ref{fig:controller}. We use an unrolled single-cell LSTM to encode any time-dependent data. The LSTM takes as input a sample from the latent space $z$ of the VAE. The hidden state of the LSTM - $h$ along with the latent sample $z$ is concatenated to form the inputs to a simple three-layered MLP (Multilayer Perceptron). The MLP outputs a $\mathrm{R}^4$ dimensional output representing the velocities and desired yaw rate of the drone. The parameters of the encoder of the VAE are fixed during training. We collect expert trajectories from the \textit{fake} environment for training. Due to the low dimensional nature of the problem, we train the network using genetic evolution. 

After this step, the drone can successfully navigate through a series of gates placed in the AirSim environment. However, we are interested in transferring this learnt policy onto an entirely different environment, where there are no gates. In essence, we want the drone to believe that it is still operating in the \textit{fake} environment with gates, but in actuality operating in a different environment without gates.

\subsection{Cheating the Network}
A block diagram of the cheating process is illustrated in fig \ref{fig:cheating}. Here, we freeze all of the weights of the network thus far and replace the encoding stage containing a different network with a pre-trained ResNet-18 \cite{resnet18} backbone. We place gates in the new environment and train the encoding stage in a supervised manner. Using this method, the new encoder network attempts to cheat the controller, and hence the drone, into performing actions without ever experiencing the actual environment. We use high fidelity indoor environments provided by AirSim and PEDRA \cite{pedra} to train the encoder network. The transferred policy can be executed on all indoor environments available on PEDRA, even if the drone has not explored any of them prior to testing.

\section{Preliminary Results}

\begin{figure}[h!]
    \centering
    \includegraphics[width = \linewidth]{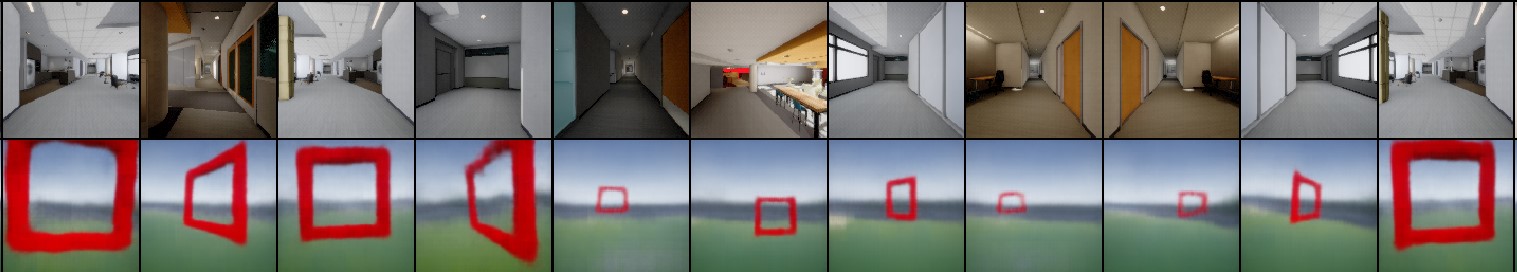}
    \caption{Cheated network - RGB inputs from \textit{real} environment (top) vs predicted state in \textit{fake} environment (bottom). The bottom row indicates what the drone \textit{believes to observe} when it is executing the policy in an unexplored environment}
    \label{fig:cheatimg}
\end{figure}

\begin{table}[ht!]
    \centering
        \begin{tabular}{SS} \toprule
        {Method} & {Mean Distance before crash (m)}\\ \midrule
        {Ours}  &   14.86\\
        {Regression} & 15.43 \\
        \bottomrule
    \end{tabular}
    \caption{Experimental results}
    \label{tab:results}
\end{table}

To experimentally validate our framework we provide some preliminary results by training the original VAE on the \textit{fake} environment generated from AirSim, where a control policy for controlling the steering angle was learnt via behavioural cloning. We then transferred this learnt policy into the Building\_99 environment on AirSim.
We compare our framework with a policy learnt directly via steering regression from RGB images using a pre-trained ResNet-18 backbone. We compare the mean distance to crash to determine whether the policy is transferred appropriately. Table \ref{tab:results} shows that our method is comparable to a policy trained directly on the indoor environment, without ever interfacing with that environment.

\section{Related Work}


Classical techniques utilize visual-inertial odometry for state estimation and model-based path planning for control of the quadcopter. World Models \cite{worldmodels} explores learning separate networks
for the environment representation and controls, instead of the end-to-end paradigm. \cite{bonatti2019learning} uses a fully learning-based pipeline for drone navigation by introducing an intermediate learnt representation for perception. 
\cite{lin2019flying} replaced planning and control with corresponding networks to navigate through a narrow gap. \cite{learningFlying} have proposed a general shallow network to replace the traditional "map-localize-plan" approach that generalizes well for indoor corridors and parking lots. \cite{learningCrashing} have used a novel approach of negative learning for navigating drones in a cluttered environment.  
Sim-to-real papers \cite{sim2real1}, \cite{sim2real2} uses popular deep learning technique of transfer learning \cite{transfer_learning_base} to transfer knowledge acquired in simulation to real world. 
 \cite{transfer_learning_base}

Although these methods attempt to solve the problem of end-to-end autonomous drone navigation, to the best of our knowledge, our work is the first to present a cheating based methodology for the problem by transferring learnt policies through different domains.

\section{Conclusion and Outlook}

\begin{figure}[h!]
    \centering
    \includegraphics[width = \linewidth]{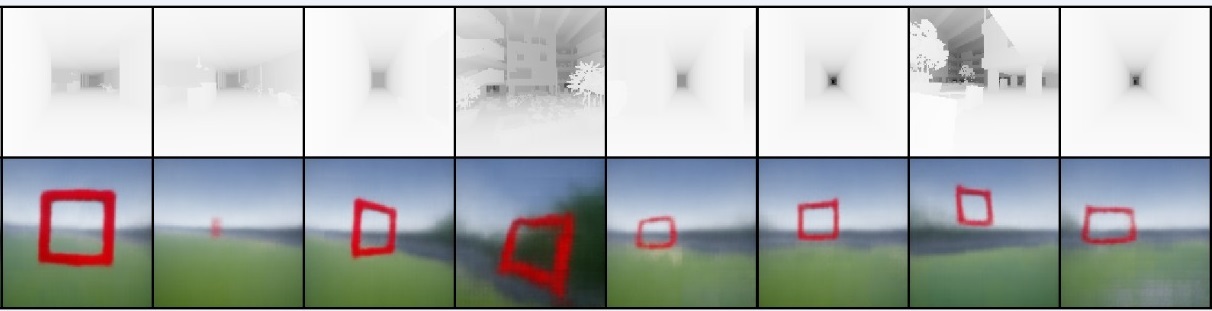}
    \caption{Cheated network in a more complex case - Depth map inputs (top) vs predicted state in \textit{fake} environment (bottom).}
    \label{fig:cheatdepth}
\end{figure}
In this paper, we present a framework that aims to reuse learnt policies to act in different environments. By cheating the learnt policy into believing that it exists in an imagined world, the learnt policy can be directly transferred onto an unseen environment. The strength of this method is twofold - 1) Ability to transfer low-level control policies to more complex problems \& 2) Ability to act as an interpretation tool (such as in Fig \ref{fig:cheatdepth} and Fig \ref{fig:cheatimg}).

For example, if we consider the toy problem of the popular CarRacing-v0 environment by OpenAI Gym \cite{openAI}, a policy learnt on this network can be transferred directly to a more complex situation such as the CARLA \cite{carla} environment using the method presented in this paper. Moreover, in its converse scenario, it is possible to transfer a learnt policy on the CARLA environment to the simpler CarRacing-v0 environment in order to interpret and visualize the policy via the decoding stage.

The framework presented in this paper can also be extended to work as a semi-guided navigation system for drones. For example, by conditioning the VAE on a 3D intent vector, we can direct where to place a particular gate in 3D space. This intent vector can be computed based on a map as a local guiding vector in a global space. The method can also be adapted to function in tandem with a text-based guidance system, where guiding intent vectors are generated from visual-textual based inputs via captioning networks.

\newpage
\bibliography{example.bib}
	
\end{document}